\documentclass[10pt,twocolumn,letterpaper]{article}

\usepackage{cvpr}
\usepackage{times}
\usepackage{epsfig}
\usepackage{graphicx}
\usepackage{amsmath}
\usepackage{amssymb}
\usepackage{xcolor}

\usepackage{multirow}
\usepackage{color}
\usepackage{comment}
\usepackage{stackengine}
\usepackage{makecell}

\usepackage[breaklinks=true,bookmarks=false]{hyperref}

\cvprfinalcopy 

\def\cvprPaperID{****} 

\ifcvprfinal\fi
\begin{document}

\title{\vspace{-0.1in}PI-Net: A Deep Learning Approach to Extract Topological Persistence Images}

\author{Anirudh Som$^{1,2}$, Hongjun Choi$^{1,2}$, Karthikeyan Natesan Ramamurthy$^3$, Matthew P. Buman$^4$, \\Pavan Turaga$^{1,2}$\\ \\
$^{1}$School of Arts, Media and Engineering, Arizona State University\\
$^{2}$School of Electrical, Computer and Energy Engineering, Arizona State University\\
$^{3}$IBM Research\\
$^{4}$College of Health Solutions, Arizona State University\\
{\tt\small asom2@asu.edu, hchoi71@asu.edu, knatesa@us.ibm.com, mbuman@asu.edu, pturaga@asu.edu}
}

\maketitle

\begin{abstract}
   Topological features such as persistence diagrams and their functional approximations like persistence images (PIs) have been showing substantial promise for machine learning and computer vision applications. This is greatly attributed to the robustness topological representations provide against different types of physical nuisance variables seen in real-world data, such as view-point, illumination, and more. However, key bottlenecks to their large scale adoption are computational expenditure and difficulty incorporating them in a differentiable architecture. We take an important step in this paper to mitigate these bottlenecks by proposing a novel one-step approach to generate PIs directly from the input data. We design two separate convolutional neural network architectures, one designed to take in multi-variate time series signals as input and another that accepts multi-channel images as input. We call these networks Signal PI-Net and Image PI-Net respectively. To the best of our knowledge, we are the first to propose the use of deep learning for computing topological features directly from data. We explore the use of the proposed PI-Net architectures on two applications: human activity recognition using tri-axial accelerometer sensor data and image classification. We demonstrate the ease of fusion of PIs in supervised deep learning architectures and speed up of several orders of magnitude for extracting PIs from data. Our code is available at \url{https://github.com/anirudhsom/PI-Net}.
\end{abstract}
\let\thefootnote\relax\footnotetext{This work was supported in part by NIH R01GM135927 and NSF CAREER 1452163. Arizona State University’s institutional review board approved all study materials and procedures (protocol number 1304009121).}

\vspace{-0.15in}
\section{Introduction}

Deep learning over the past decade has had tremendous impact in computer vision, natural language processing, machine learning, and healthcare. 
Among other approaches, convolutional neural networks (CNNs) in particular have received great attention and interest from the computer vision community. This is attributed to the fact that they are able to exploit the local temporal and spatial correlations that exist in 1-dimensional (1D) sequential time-series signals, 2-dimensional (2D) data like images, 3-dimensional (3D) data like videos, and 3D objects. In this paper, we refer to these type of data as \textit{input data}. 
CNNs also have far less learnable parameters than their fully-connected counterparts, making them less prone to over-fitting 
and have shown state-of-the-art results in applications like image classification, object detection, scene recognition, fine-grained categorization and action recognition \cite{krizhevsky2012imagenet,girshick2014rich,zhang2014part,zhang2014panda,zhou2014learning}.
Apart from being good at learning mappings between the input and corresponding class labels, deep learning frameworks are also efficient in discovering mappings between the input data and other output feature representations \cite{walker2015dense,wang2015designing,long2015fully,eigen2014depth,dong2014learning}. 


\begin{figure}[t!]
	\centering
	\includegraphics[width=1\linewidth]{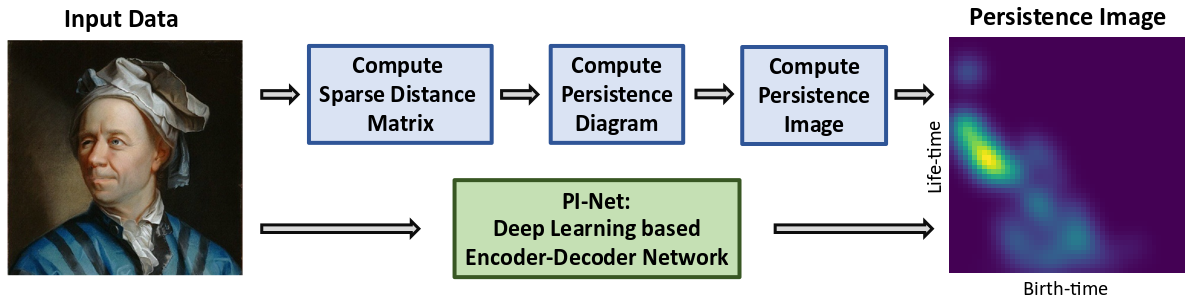}
	\caption{Illustration of the proposed PI-Net model to directly compute persistence images from input data. Traditional analytic methods (illustrated in the top half of the figure) consist of a sequence of steps that are computationally expensive.}\label{main_idea_illustration}
	\vspace{-0.2in}
\end{figure}

While methods for learning features from scratch and mapping image data to desired outputs via neural networks have matured significantly, relatively less attention has been paid to invariance to nuisance low-level physical factors like sensor noise. 
Topological data analysis (TDA) methods are popularly used to characterize the \textit{shape} of $n$-dimensional point cloud data using representations such as persistent diagrams (PDs) that are robust to certain types of variations in the data \cite{edelsbrunner2010computational}. TDA methods have also been successfully applied to different computer vision problems and have shown the ability to incorporate different invariances of interest to the computer vision community \cite{li2014persistence,Dey2017,som2018perturbation}. The shape of the data is quantified by properties such as connected components, cycles, high-dimensional holes, level-sets and monotonic regions of functions defined on the data \cite{edelsbrunner2010computational}. Topological properties are those invariants that do not change under smooth deformations like stretching, bending and rotation, but without tearing or gluing surfaces. These attractive traits of TDA have renewed interested in this area for answering various fundamental questions, including those dealing with interpretation, generalization, model selection, stability, and convergence \cite{gabrielsson2018exposition,bubenik2016statistical,rieck2018neural,ramamurthy2018topological,gabella2019topology,ferri2018topology}. 

A lot of work has gone into utilizing topological representations efficiently in large-scale machine learning \cite{anirudh2016riemannian,bubenik2015statistical,rouse2015feature,pachauri2011topology,reininghaus2015stable,adams2017persistence,som2018perturbation}. However, bottlenecks such as computational load involved in discovering topological invariants as well as a lack of a differentiable architecture remain. In this paper we propose simple deep learning architectures to learn approximate mappings between data and their topological feature representations.
The gist of our idea is illustrated in Figure \ref{main_idea_illustration} and the main contributions are listed below. 

\noindent \textbf{Contributions:} 
\vspace{-0.075in}
\begin{enumerate}
    \item We propose a novel differentiable neural network architecture called \emph{PI-Net}, to extract topological representations. In this paper we focus on persistence images (PIs) as the desired topological feature.
    \vspace{-0.1in}
    \item We provide two simple CNN-based architectures called \emph{Signal PI-Net} that takes in multi-variate 1D sequential data and \emph{Image PI-Net} that takes in multi-channel 2D image data.
    \vspace{-0.1in}
    \item We explore transfer learning strategies to train the proposed \emph{PI-Net} model on a source dataset and use it on a different target dataset, with or without fine-tuning.
    \vspace{-0.1in}
    \item Through our experiments on human activity recognition using accelerometer sensor data and image classification on standard image datasets, we show the effectiveness of the generated approximations for PIs and compare their performance to PIs generated using analytic TDA methods. We also investigate the benefits of concatenating PIs with features learnt using deep learning methods. 
    \vspace{-0.1in}
    \item We also evaluate the robustness of classification models to Gaussian noise, with or without fusion with PI representations in image classification tasks.
\end{enumerate}

The rest of the paper is outlined as follows: Section \ref{related_work} discusses related work. Section \ref{background_conceps} provides the necessary background on TDA, PIs and CNNs. In Section \ref{proposed_method} we describe the proposed \emph{PI-Net} frameworks in detail and in Section \ref{experimental_results} we describe the experimental results. Section \ref{conclusion_future_work} concludes the paper.

\vspace{-0.05in}
\section{Related Work}\label{related_work}

Although the formal beginnings of topology is already a few centuries old dating back to Euler, algebraic topology has seen a revival in the past decade with the advent of computational tools and software \cite{scikittda2019,adams2014javaplex,bauer2014distributed}. Arguably the most popular topological summary is the persistence diagram (PD), which is a multi-set of points in a 2D plane that quantifies the \textit{birth} and \textit{death} times of topological features such as $k$-dimensional holes or sub-level sets of a function defined on a point cloud \cite{Edelsbrunner2002}. This simple summary has resulted in the adoption of topological methods for various applications \cite{perea2015sliding,tralie2018quasi,chintakunta2015entropy,dabaghian2012topological,chung2009persistence,heath2010image,singh2008topological,venkataraman2016persistent,nawar2020topological}. 
However, TDA methods suffer from two major limitations. First, it is computationally very taxing to extract PDs. The computational load increases both with the dimensionality and with the number of samples in the data being analyzed. The second obstacle is that a PD is a multi-set of points, making it impossible to use machine learning or deep learning frameworks directly on the space of PDs. Efforts have been made to tackle the second issue by attempting to map PDs to spaces that are more favorable for machine learning tools \cite{anirudh2016riemannian,bubenik2015statistical,rouse2015feature,pachauri2011topology,reininghaus2015stable,adams2017persistence,som2018perturbation}. For further reading, \cite{somgeometric} surveys recent topological representations and their associated metrics. To alleviate the first problem, in this paper we propose a simple one-step differentiable architecture called \emph{PI-Net} to compute the desired topological feature representation, specifically persistence images. To the best of our knowledge, we are the first to propose the use of deep learning for computing PIs directly from data.

Our motivation to use deep learning stems from its successful use to learn mappings between input data and different feature representations \cite{walker2015dense,wang2015designing,long2015fully,eigen2014depth,dong2014learning}. However, deep learning and TDA did cross paths before but not in the same context as what we propose in this paper. TDA methods have been used to study the topology \cite{gabrielsson2018exposition,bubenik2016statistical}, algorithmic complexity \cite{rieck2018neural}, behavior \cite{gabella2019topology} and selection \cite{ramamurthy2018topological} of deep learning models. Efforts have also been made to use topological feature representations either as inputs or fused with features learned using neural network models \cite{Dey2017,hofer2017deep,cang2017topologynet}. Later in Section \ref{experimental_results}, we show experimental results on fusing generated PIs with deep learning frameworks for action recognition and image classification.


\vspace{-0.05in}
\section{Background}\label{background_conceps}

\begin{figure}[t!]
	\centering
	\scalebox{0.9}{
	\includegraphics[width=1\linewidth]{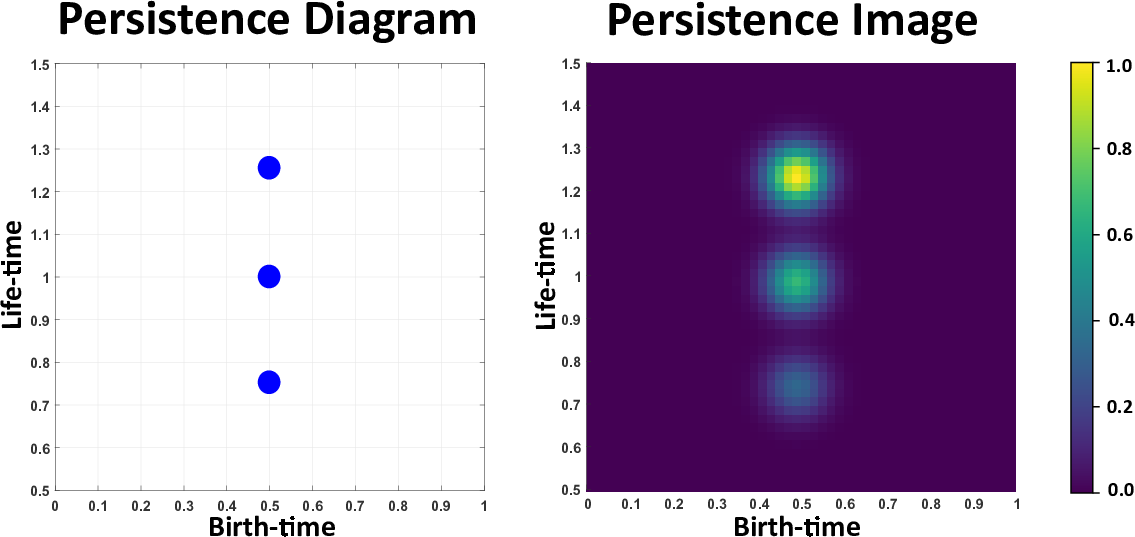}}
	\caption{Illustration of a PD and its weighted PI for three points with same birth-time but different life-time. Due to the weighting function points with higher life-time appear more brighter.}\label{pd_pi_fig}
	\vspace{-0.225in}
\end{figure}

\noindent \textbf{Persistence Diagrams:}  Consider a graph $\mathcal{G} = \{\mathcal{V}, \mathcal{E}\}$ constructed from data projected onto a high-dimensional point-cloud space. Here, $\mathcal{V}$ is the set of $|\mathcal{V}|$ nodes and $\mathcal{E}$ denotes the neighborhood relations defined between the samples. Topological properties of the graph can be estimated by first constructing a simplicial complex $\mathcal{S}$ over $\mathcal{G}$. $\mathcal{S}$ is defined as $\mathcal{S}=(\mathcal{G},\Sigma)$, with $\Sigma$ being a family of non-empty level sets of $\mathcal{G}$, with each element $\sigma \in \Sigma$ is a simplex \cite{Edelsbrunner2002}. This falls under the realm of \emph{persistent homology} where we are interested in summarizing the $k$-dimensional holes present in the data. The simplices are constructed using the the $\epsilon$-neighborhood rule \cite{Edelsbrunner2002}. It is also possible to quantify the topology induced by a function $g$ defined on the vertices of a graph $\mathcal{G}$ by studying the topology of its sub-level or super-level sets. Since $g: \mathcal{V} \rightarrow \mathbb{R}$, this is referred to as \emph{scalar field topology}. In either case, PDs provide a simple way to summarize the birth vs death time information of the topological feature of interest. In this paper we use \emph{persistent homology} to compute ground-truth PDs for images and \emph{scalar field topology} to compute ground-truth PDs for time-series signals. In a PD the birth-time $b$ refers to the scale at which the feature was formed and death-time $d$ refers to the scale at which it ceases to exist. The difference between $d$ and $b$ gives us the life-time or persistence and is denoted by $l = |d-b|$. Each PD is a multi-set of points $(b,d)$ in $\mathbb{R}^2$. Since $d\geq b$, only one-half of the space in the PD is actually utilized. Points in the PD that lie close to the diagonal represent noise and can be easily discarded by simple thresholding. Plotting the birth-time vs life-time information allows us to utilize the entire 2D space of a PD as shown in Figure \ref{pd_pi_fig}. Interested readers can refer to the following papers to learn more about the properties of the space of PDs \cite{edelsbrunner2010computational,Edelsbrunner2002}. 

\noindent \textbf{Persistence Images:} A PI is a finite-dimensional vector representation of a PD \cite{adams2017persistence} and can be computed through the following series of steps. First we map the PD to an integrable function $\rho: \mathbb{R}\rightarrow\mathbb{R}^2$ called a persistence surface. The persistence surface $\rho$ is defined as a weighted sum of Gaussian functions that are centered at each point in the PD. Next, a discretization of a sub-domain of the persistence surface is done which results in a grid. Finally, the PI is obtained by integrating the persistence surface over each grid box, giving us a matrix of pixel values. An interesting aspect when computing PIs is the broad range of weighting functions to chose from, to weight the Gaussian functions. Typically, points of high persistence or lifetime are perceived to be more important than points of low persistence. In such cases one may select the weighting function to be non-decreasing with respect to the persistence value of each point in the PD. Adams \textit{et al.} also talk about the stability of persistence images with respect to the 1-Wasserstein distance between PDs \cite{adams2017persistence}. Figure \ref{pd_pi_fig} illustrates an example of a PD and its PI where the points are weighted by their life-time.

\noindent \textbf{Convolutional Neural Networks:} CNNs were inspired from the hierarchical organization of the human visual cortex \cite{grill2004human} and consist of many intricately interconnected layers of neuron structures serving as the basic units to learn, extract both low-level and high-level features from images. CNNs are particularly more attractive and powerful compared to their connected counterparts because CNNs are able to exploit the spatial correlations present in natural images and each convolutional layer has far less trainable parameters than a fully-connected layer. Several sophisticated CNN architectures have been proposed in the last decade, for example \emph{AlexNet} \cite{krizhevsky2012imagenet}, \emph{VGG} \cite{simonyan2014very}, \emph{GoogleNet} \cite{szegedy2015going}, \emph{ResNet} \cite{he2016deep}, \emph{DenseNet} \cite{huang2017densely}, \emph{etc}. Some of these designs are known to surpass humans for object recognition tasks \cite{silver2016mastering}. Apart from discovering features from scratch for classification tasks, CNNs are also popular for learning mappings between input and other feature representations \cite{walker2015dense,wang2015designing,long2015fully,eigen2014depth,dong2014learning}. This motivates us to design simple CNN models for the task of learning mappings between the data and their PI representations. We would like to direct interested readers to the following survey paper to know more about different CNN architectures \cite{liu2017survey,srinivas2016taxonomy}.

\noindent \textbf{Learning Strategies:} Here we will briefly talk about the two learning strategies namely: \emph{Supervised Learning} and \emph{Transfer Learning}. We employ these strategies to train the proposed \emph{PI-Net} model. \emph{Supervised Learning} is concerned with learning complex mappings from $X$ to $Y$ when many pairs of $(x,y)$ are given as training data, with $x\in X$ being the input data and $y\in Y$ being the corresponding label or feature representation. In a classification setting $Y$ corresponds to a fixed set of labels. In a regression setting, the output $Y$ is either a real number or a set of real numbers. In this paper our problem falls under the regression category as we try to learn a mapping between the input data and its PI. 
\emph{Transfer Learning} is a design methodology that involves using the learned weights of a pre-trained model that is trained on a source dataset $\mathcal{D}_s$ for the source task $\mathcal{T}_s$, to initialize the weights of another model that is fine-tuned using a target dataset $\mathcal{D}_t$ for the target task $\mathcal{T}_t$ \cite{yosinski2014transferable}. This allows us to leverage the source dataset that the model was initially trained on without having to train the model from scratch. The is useful in cases where the target dataset has a lot less data samples compared to the source dataset. In Section \ref{proposed_method} we show how transfer learning is employed in our proposed framework when the target training data is limited.


\begin{figure*}[t!]
	\centering
	\includegraphics[width=0.9\linewidth]{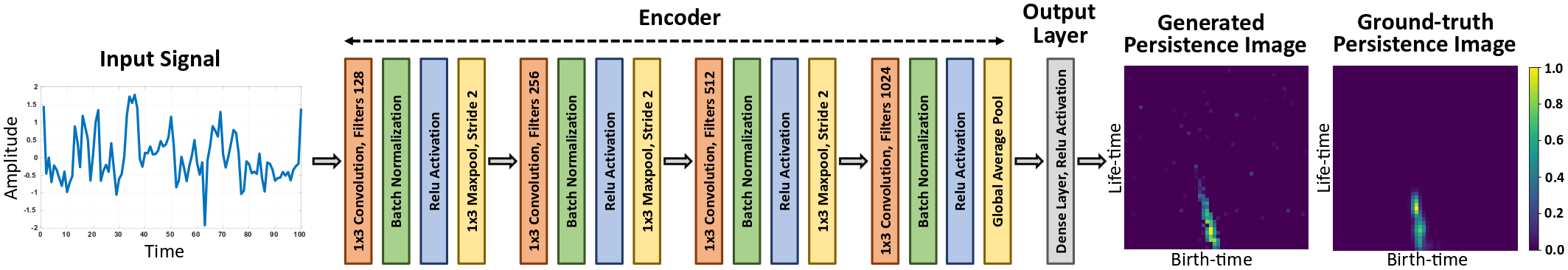}
	\vspace{-0.025in}
	\caption{Illustration of Signal PI-Net for computing PIs directly from multi-variate time-series signals.}\label{signal_pinet_illustration}
	\vspace{-0.1in}
\end{figure*}

\begin{figure*}[t!]
	\centering
	\includegraphics[width=0.9\linewidth]{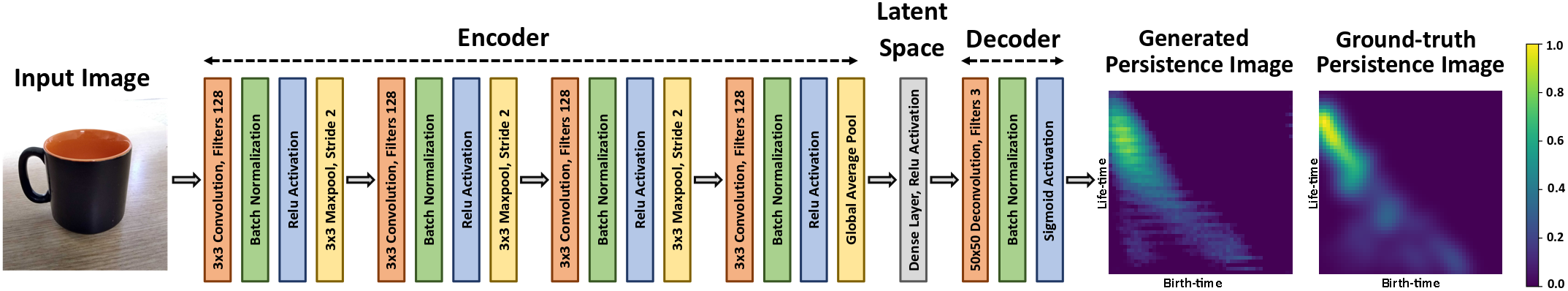}
	\vspace{-0.025in}
	\caption{Illustration of Image PI-Net for computing PIs directly from multi-channel image data.}\label{image_pinet_illustration}
	\vspace{-0.1in}
\end{figure*}

\vspace{-0.05in}
\section{PI-Net Framework}\label{proposed_method}

In this section we first describe the steps to generate ground-truth PIs and later discuss the proposed \emph{Signal PI-Net} and \emph{Image PI-Net} configurations. 

\subsection{Generating Ground-truth Persistence Images}

\noindent \textbf{Data Pre-processing:} For uni-variate or multi-variate time-series signals, we consider only fixed-frame signals, \emph{i.e.} signals with fixed number of time-steps, and zero-center them. We standardize the training and test sets such that they have unit variance along each time-step. For images we enforce the pixel values to range between $[0, 1]$.

\noindent \textbf{Persistence Images for Time-series Data:}  We use the \emph{Scikit-TDA} python library \cite{scikittda2019} and use the \emph{Ripser} package for computing PDs. As mentioned earlier, we compute \emph{level-set} filtration PDs for time-series signals. 
\emph{Scalar field topology} offers a simple way to summarize the different peaks and troughs present in the signal. For example a local minima gives birth to a topological feature (more accurately a 0-dimensional homology group feature) which dies at its local maxima. We compute PDs for each of the $x, y, z$ signals in the accelerometer sample. For better use of the 2D space in the PD we consider birth-time vs life-time information. For computing PIs we used the \emph{Persim} package in the \emph{Scikit-TDA} toolbox. In all our experiments we set the grid size of the generated PIs to 50$\times$50 and fit a Gaussian kernel function on each point in the PD. We weight each Gaussian kernel by the life-time of the point. For all time-series datasets we set the standard deviation of the Gaussian kernel to 0.25 and set the birth-time range to [-10, 10]. Once computed we normalize each PI by dividing by its maximum intensity value. This forces the values in the PI to also lie between [0,1]. 

\noindent \textbf{Persistence Images for Multi-channel Image Data:} Here too we use the \emph{Scikit-TDA} library for computing PDs and PIs. We represent each image channel as a 3D point cloud with the three coordinates representing the $x$-coordinate, $y$-coordinate and intensity value of each pixel in the image. For example, an image with $c$ channels will result in $c$ 3D point clouds. The $x$ and $y$ coordinate information is also normalized to be within $[0, 1]$. Finally, we compute the $1$-dimensional persistent homology PDs for each channel in the image using the process described in Section \ref{background_conceps}. For all image datasets in our experiments we discard points in the PD with life-time less than $0.02$. For computing PIs we set the grid size of the generated PIs to 50$\times$50 and fit a Gaussian kernel function on each point in the PD. The Gaussian kernel is weighted by the life-time of the point. Other parameters needed to compute PIs like birth-time range and standard-deviation of the Gaussian kernel were set to different values specific to each dataset. We consider the following three datasets in our experiments: \emph{CIFAR10} \cite{krizhevsky2009learning}, \emph{CIFAR100} \cite{krizhevsky2009learning} and \emph{SVHN} \cite{netzer2011reading}. For \emph{CIFAR10} and \emph{CIFAR100} we set birth-time range and standard-deviation to $[0, 0.3]$ and $0.01$. For \emph{SVHN} we set the same parameters to $[0, 0.2]$ and $0.005$ respectively. Finally, each of the $c$ PIs generated for a $c$-channel image is further normalized to lie in the range $[0, 1]$.

\subsection{Network Architecture} Both \emph{PI-Net} architectures were designed using Keras with TensorFlow back-end \cite{chollet2015keras}.

\noindent \textbf{Signal PI-Net:} The input to the network is a $b\times t\times n$ dimensional time-series signal, where $b$ is the batch-size, $t$ refers to the number of time-steps or frame size. For a uni-variate signal $n = 1$ and for a multi-variate signal $n>1$. For our experiments in section \ref{experimental_results}, $n$ is $3$ and $t$ is either $250$ or $500$. After the input layer, the encoder block consists of four 1D convolution layers. Except the final convolution layer, all other convolution layers are followed by batch normalization, ReLU activation and Max-pooling. The final convolution layer is followed by batch normalization, ReLU activation and Global-average-pooling. The number of convolution filters is set to 128, 256, 512 and 1024 respectively. However, the convolution kernel size is same for all layers and is set to 3 with stride set to 1. We use appropriate zero padding to keep the output shape of the convolution layer unchanged. For all Max-pool layers, we set the kernel size to 3 and stride to 2. After the encoder block, we pass the global-average-pooled output into a final output dense layer of size $2500 \times n$. The output of the dense layer is subjected to ReLU activation and reshaped to size $50\times50\times n$. 

\begin{figure}[t!]
	\centering
	\includegraphics[width=0.95\linewidth]{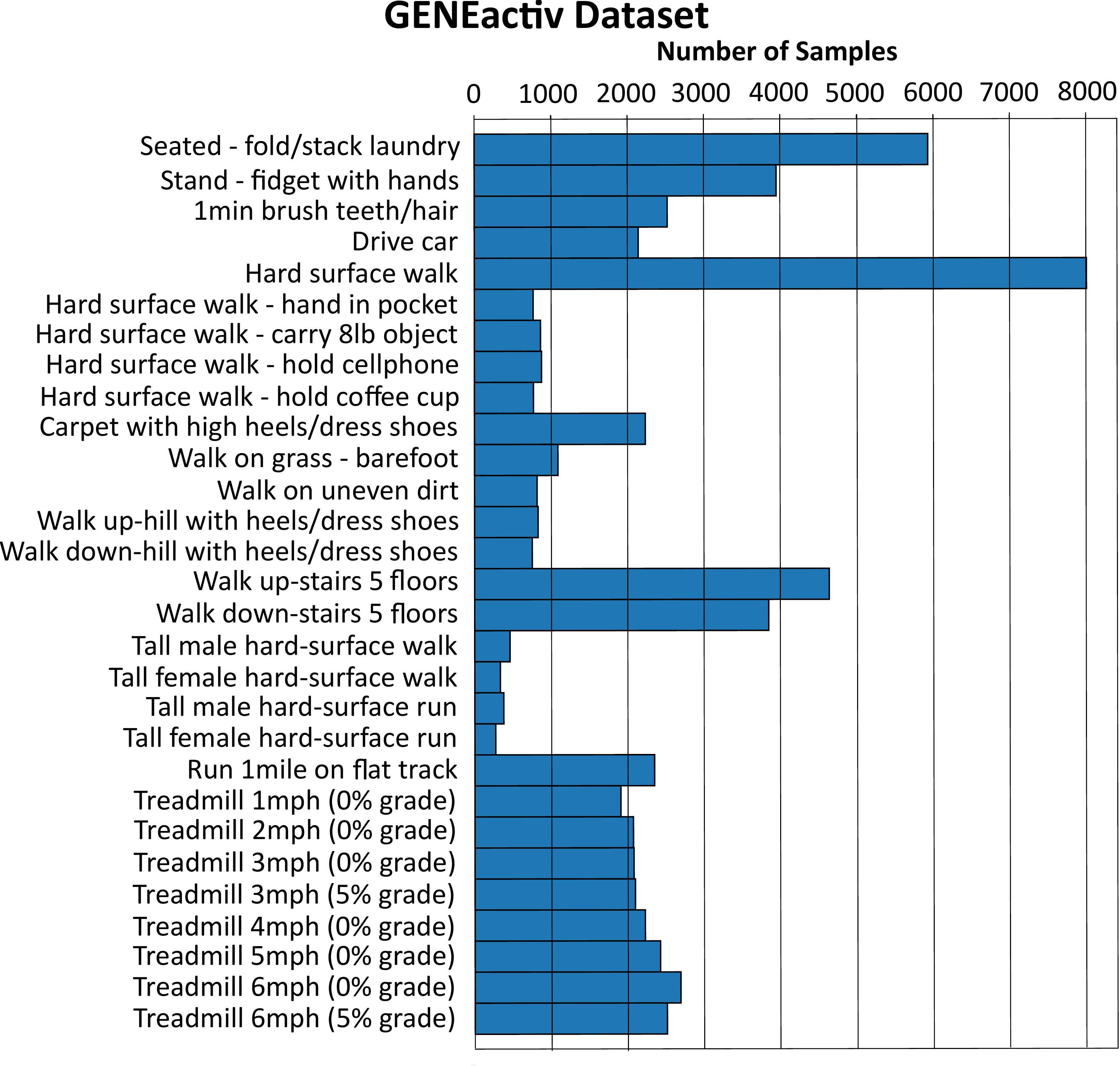}
	\caption{Distribution of activity classes in the \emph{GENEactiv} dataset for time-step length = 250.} \label{geneactiv_distribution}
	\vspace{-0.25in}
\end{figure}

\noindent \textbf{Image PI-Net:} The input to this network is a $b\times h\times w \times c$ dimensional image, where $b, h, w, c$ is the batch-size, the image height, width and number of channels respectively. \emph{Image PI-Net} follows the same architecture as \emph{Signal PI-Net} for the encoder block. However, we now use 2D convolution layers instead. Also, for all the convolution layers the number of filters and kernel size  was set to 128 and 3 respectively. We use appropriate zero padding to keep the output shape of the convolution layer unchanged. For all Max-pool layers, we set the kernel size to 3 and stride to 2. We pass the output of the encoder block into a latent variable layer which consists of a dense layer of size 2500. The output of the latent variable layer is reshaped to $50\times50$ and is passed into the decoder block. The decoder block consists of one 2D deconvolution layer with kernel size set to 50, stride set to 1, number of filters to $c$. The output of the deconvolution layer is also zero-padded such that the height and width of the output remain unchanged. The deconvolution layer is followed by a final batch normalization and Sigmoid activation. The shape of the output we get is  $50\times50\times c$.

\vspace{-0.05in}
\section{Experiments}\label{experimental_results}

This section can be broadly divided into four parts. First we show results for human activity recognition by using PIs alone and PIs in fusion with different deep learning models on two accelerometer sensor datasets: \emph{GENEactiv} \cite{wang2016statistical} and \emph{USC-HAD} \cite{zhang2012usc}. Second, we show image classification results with and without fusing PIs with a \emph{DenseNet} \cite{huang2017densely} classifier on the following image datasets: \emph{CIFAR10} \cite{krizhevsky2009learning} and \emph{SVHN} \cite{netzer2011reading}. Third, we show how the generated PIs together with the  image classification model can help improve robustness to \emph{Gaussian noise}. Finally, we show improvements in computation time for the task of extracting PIs from image databases using \emph{Image PI-Net}.

\vspace{-0.05in}
\subsection{Action Recognition using Accelerometer Data}\label{accelerometer_expts}
\vspace{-0.05in}

\begin{figure}[t!]
	\centering
	\includegraphics[width=0.85\linewidth]{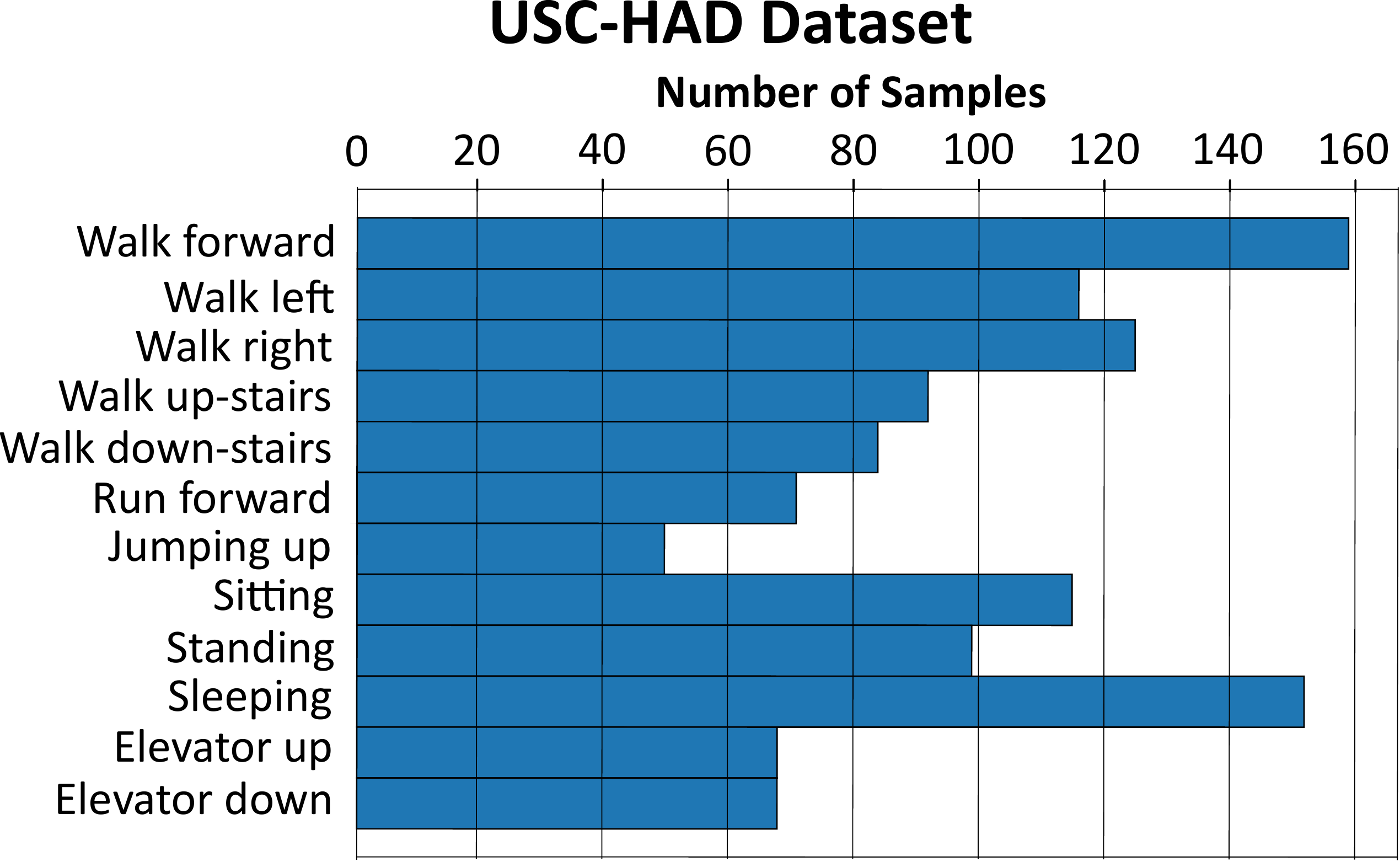}
	\caption{Distribution of activity classes in the \emph{USC-HAD} dataset for time-step length = 250.} \label{uschad_distribution}
	\vspace{-0.15in}
\end{figure}

\noindent\textbf{Dataset Description:} The \emph{GENEactiv} dataset consists of 29 different human-activity classes from 152 subjects \cite{wang2016statistical}. The data was collected at a sampling rate of 100Hz using the \emph{GENEactiv} sensor, a light-weight, waterproof, wrist-worn tri-axial accelerometer. Interested readers can refer to the following paper to learn about the data collection protocol \cite{wang2016statistical}. The \emph{USC-HAD} dataset consists of 12 different human-activity classes from 14 subjects \cite{zhang2012usc}. Data was collected using a tri-axial \emph{MotionNode} accelerometer sensor at a sampling rate of 100Hz. The sensor was placed at the front right hip on the body. Both datasets were down-sampled to 50Hz and fixed-length non-overlapping frames were extracted. Figures \ref{geneactiv_distribution} and \ref{uschad_distribution} show the distribution of the different activity classes in each dataset, with each frame having a duration of 5 seconds or 250 time-steps. For the \emph{GENEactiv} dataset we extracted frames with time-steps = 250 and 500, and used approximately 75\% of the frames for training and the rest as the test set. \emph{USC-HAD} being a significantly smaller dataset, we only extracted frames with time-step = 250 and used frames from the first 8 subjects for training and the remaining 6 subjects as the test set.

\noindent \textbf{Training Signal PI-Net:} The \emph{Signal PI-Net} model was trained using just the training set of the \emph{GENEactiv} dataset. The batch-size was set to 128 and the model was trained for a 1000 epochs. The learning rate for the first 300 epochs, next 300 epochs and final 400 epochs was set to $10^{-3}$, $10^{-4}$ and $10^{-5}$ respectively. \emph{Adam} optimizer was used and the \emph{Mean-Squared-Error} loss function was used to quantify the deviation of the generated PIs from the ground-truth PIs. Final training and test loss values are tabulated in Table \ref{train_test_loss_values}. 

\begin{table}[t!]
	\centering
	\scalebox{0.85}{\begin{tabular}{ |c|c|c|c| } 
			\hline
			\textbf{PI-Net} & \textbf{Train Loss}  & \textbf{Test Loss} \\
			\hline
			\makecell{Signal PI-Net\\ Time-steps = 250} & 0.00159  & 0.00158 \\
			\makecell{Signal PI-Net\\ Time-steps = 500} & 0.00187 & 0.00187 \\
			\hline
			CIFAR10 Image PI-Net & 1.99793 & 2.06193 \\
			CIFAR10 Image PI-Net FA & 2.02095 & 2.04441 \\
			CIFAR10 Image PI-Net FS & 0.51734 & 0.52560 \\
			\hline
			SVHN Image PI-Net & 1.53533 & 1.51732 \\
			SVHN Image PI-Net FA & 1.57923 & 1.54195 \\
			SVHN Image PI-Net FS & 0.41519 & 0.40955 \\
			\hline
	\end{tabular}}
	\vspace{0.05in}
	\caption{Final train and test loss values after training the different Signal PI-Net and Image PI-Net models.}\label{train_test_loss_values}
	\vspace{-0.3in}.
\end{table}

\noindent\textbf{Data Characterization and Classification:} For characterizing the time-series signals, we consider three different feature representations: \textbf{(1)} A 19-dimensional feature vector consisting of different statistics calculated over each 10-second frame \cite{wang2016statistical}; \textbf{(2)} Features learnt from scratch using multi-layer-perceptron (MLP) models and 1D CNNs; \textbf{(3)} Persistence Images generated using the traditional filtration technique and the proposed \emph{Signal PI-Net} model. The 19-dimensional feature vector includes \emph{mean}, \emph{variance}, \emph{root-mean-square} value of the raw accelerations on each of $X$, $Y$ and $Z$ axes, \emph{pearson correlation coefficients} between $X$-$Y$, $Y$-$Z$ and $X$-$Z$ time series, \emph{difference between maximum and minimum accelerations} on each axis denoted by $dx, dy, dz$, and $\sqrt{dx^2 + dy^2}$, $\sqrt{dy^2 + dz^2}$, $\sqrt{dx^2 + dz^2}$, $\sqrt{dx^2 + dy^2 + dz^2}$. From here on out we will refer to this 19-dimensional statistical feature as SF. 

The MLP classifier contains 8 dense layers, with each layer having 128 units and ReLU activation. To avoid over-fitting, each dense layer is followed by a dropout layer with a dropout rate of 0.2 and a batch-normalization layer. The output layer is another dense layer with Softmax activation and with number of units equal to the number of classes. The 1D CNN classifier consists of 10 CNN layers with number of filters set to 32, kernel size to 3, stride to 1 and the output is zero-padded. Each CNN layer is followed by batch-normalization, ReLU activation and max-pooling layers. For max-pool layers we set the filter size to 3, the stride was set to 1 for every odd layer and 2 for every even layer. For the final CNN layer we use a global-average-pooling layer instead of a max-pool layer. Here too, the output layer consists of a dense layer with softmax activation and number of units equal to number of target classes.

We used the trained \emph{Signal PI-Net} model to extract PIs for the test set of the \emph{GENEactiv} dataset. We also use the same model to compute PIs for both the training and test sets of the \emph{USC-HAD} dataset. The different classification methods are listed in Table \ref{time_series_classification_geneactiv}. The PIs obtained using traditional analytic methods or using the proposed \emph{Signal PI-Net} model were fused with the MLP and 1D CNN classification models differently. For instance, \emph{MLP - PI} and \emph{MLP - Signal PI-Net} use the MLP classifier to learn features directly from the computed PIs (The PIs were vectorized and passed as inputs). \emph{MLP - SF} uses the MLP classifier with the 19-dimensional statistical feature as input. In \emph{MLP - SF+PI} and \emph{MLP - SF+Signal PI-Net} we first concatenate the SF and PI representations before passing them as input to the MLP model. However, for \emph{1D CNN + PI} and \emph{1D CNN + Signal PI-Net} we use a slightly different approach. Using Principal Component Analysis (PCA) we first reduce the vectorized PI representation (7500-dimensional) to a 32-dimensional feature vector. This was done to reduce the number of additional parameters that would result from the concatenation of the PI feature representations to the \emph{1D CNN} model. The 32-dimensional PI representation is then concatenated to the output of the global-average-pool layer in the \emph{1D-CNN} model. 

\begin{table}[t!]
	\centering
	\scalebox{0.625}{\begin{tabular}{ |c|c|c|c| } 
			\hline
			\multirow{2}{*}{\textbf{Method}} & \multicolumn{2}{c|}{\textbf{GENEactiv}} & \textbf{USC-HAD}\\
			\cline{2-4} & \textbf{Time-steps = 250} & \textbf{Time-steps = 500} & \textbf{Time-steps = 250} \\
			\hline
			MLP - PI & 46.45$\pm$0.32 & 49.67$\pm$0.63 & 43.21$\pm$0.66\\
			\textbf{MLP - Signal PI-Net} &  \textbf{49.47$\pm$0.69}  &  \textbf{53.69$\pm$1.08} & \textbf{48.15$\pm$0.67}\\
			\hline
			MLP - SF & 41.70$\pm$0.41 & 42.01$\pm$0.42  & 35.68$\pm$0.11\\ 
			MLP - SF + PI & 48.57$\pm$0.37 & 49.82$\pm$0.62 & 44.31$\pm$0.36\\ 
			\textbf{MLP - SF + Signal PI-Net} & \textbf{50.66$\pm$0.78} & \textbf{54.44$\pm$0.80} & \textbf{48.97$\pm$0.30}\\ 
			\hline
			1D CNN & 53.56$\pm$0.31 & 54.97$\pm$1.35 & 54.58$\pm$0.64\\
			1D CNN + PI & 54.28$\pm$0.23 & 56.38$\pm$0.23 & 54.64$\pm$0.62\\
			\textbf{1D CNN + Signal PI-Net} & \textbf{54.41$\pm$0.21} & \textbf{56.41$\pm$0.22} & \textbf{57.82$\pm$0.78}\\
			\hline
	\end{tabular}}
	\vspace{0.05in}
	\caption{Weighted F1 score classification results for the GENEactiv and USC-HAD datasets. The mean $\pm$ std values were calculated over five runs.}\label{time_series_classification_geneactiv}
	\vspace{-0.1in}
\end{table}

The weighted F1 score classification results for \emph{GENEactiv} and \emph{USC-HAD} is shown in Table \ref{time_series_classification_geneactiv}. For each method we report the mean $\pm$ std result over 5 runs. We observe similar results under the different time-step settings in \emph{GENEactiv} and also across the two datasets. PIs computed analytically or using \emph{Signal PI-Net} perform better than SF. Fusing PIs with SF helps significantly improve the classification performance. 1D CNN is a more powerful classifier than MLP, which is made clearly evident from the tabulated results. Fusing PIs with features learnt using 1D CNNs helps marginally improve the overall classification result.

\subsection{Image Classification}\label{image_class_section}

\begin{figure}[t!]
	\centering
	\vspace{-0.12in}
	\includegraphics[width=0.99\linewidth]{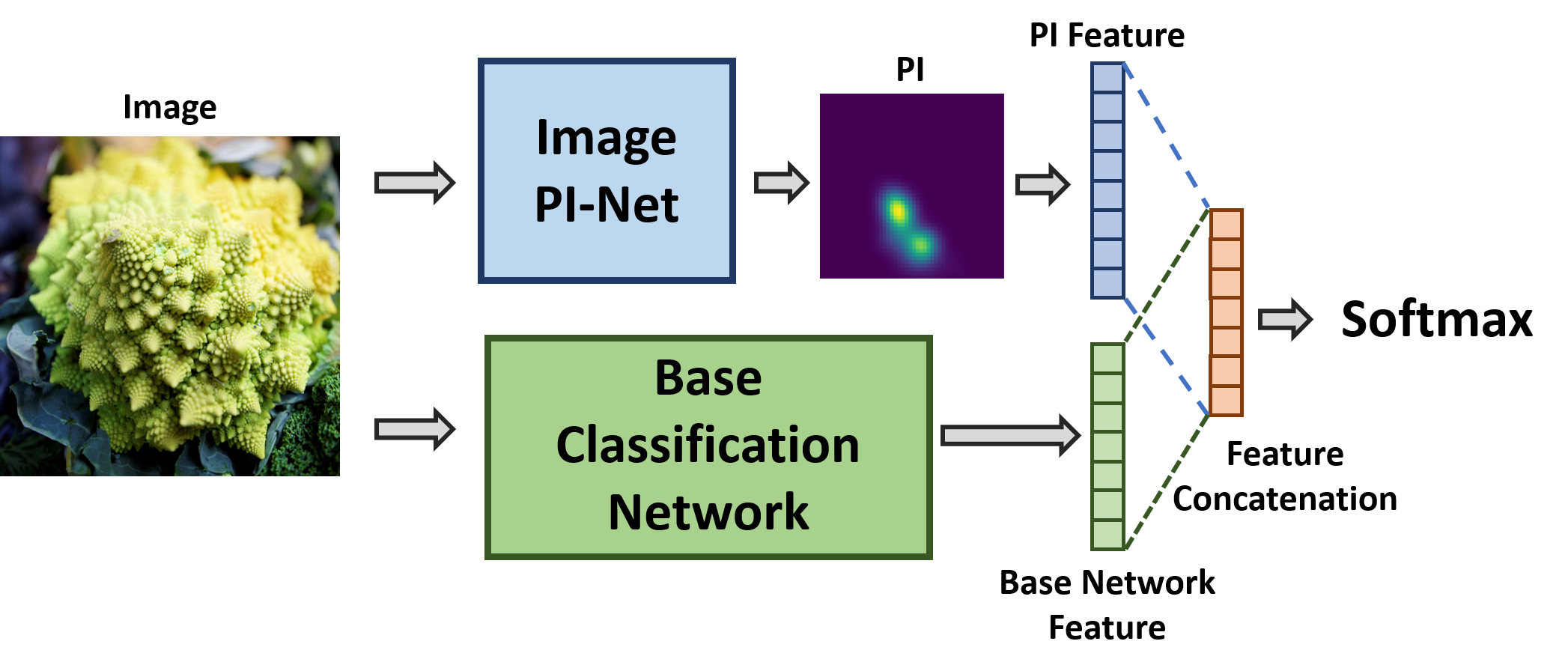}
	\vspace{-0.08in}
	\caption{Illustration of the modified base model where we concatenate PI feature with features learnt using the base classification network.}\label{fusion_illustration} 
	\vspace{-0.2in}
\end{figure}

\noindent\textbf{Dataset Description:} We consider the following three datasets in our experiments: \emph{CIFAR10} \cite{krizhevsky2009learning}, \emph{CIFAR100} \cite{krizhevsky2009learning} and \emph{SVHN} \cite{netzer2011reading}. \emph{CIFAR10} and \emph{CIFAR100} each contain 50000 images for training and 10000 images for testing, whereas \emph{SVHN} has 73257 images for training and 26032 images for testing. For classification experiments we only show results for \emph{CIFAR10} and \emph{SVHN}. Both datasets have 10 different label categories. Also, the height, width and number of channels for each image is equal to 32, 32 and 3 respectively. 

\noindent\textbf{Training Image PI-Net:} We develop two kinds of \emph{Image PI-Net} models based on the datasets we chose as source and target datasets: \textbf{(1)} In the first kind we set the source and target datasets to be same, \emph{i.e.} we train the \emph{Image PI-Net} model using the \emph{CIFAR10} or \emph{SVHN} dataset. \textbf{(2)} For the second type, we use the \emph{CIFAR100} dataset as the source dataset and the target dataset is either \emph{CIFAR10} or \emph{SVHN}. Simply put, we employ transfer learning by first training the \emph{Image PI-Net} model using \emph{CIFAR100} and later use the target dataset to fine-tune the \emph{Image PI-Net} model. For the second case, we further explore two variations: \textbf{(2a)} Fine-tune the model using all samples from the training set of the target dataset; \textbf{(2b)} fine-tune using just a subset \emph{i.e.} 500 images per class in the training set of the target dataset, to simulate the scenario of having limited training data. We will refer to these variants as \emph{Image PI-Net Fine-tune All} (\emph{Image PI-Net FA}) and \emph{Image PI-Net Fine-tune Subset} (\emph{Image PI-Net FS})  respectively. We explored the above variants to show the use of the proposed \emph{Image PI-Net} model under different scenarios. We set the batch-size to $32$. We train the basic \emph{Image PI-Net} model for 415 epochs and set the learning rate for the first 15 epochs, next 200 epochs and final 200 epochs to $10^{-3}$, $10^{-5}$ and $10^{-6}$ respectively. For \emph{Image PI-Net FA} and \emph{Image PI-Net FS} we first load the weights from the \emph{CIFAR100} pre-trained model and fine-tune the weights for 200 epochs with a learning rate of $10^{-6}$. We use the \emph{Adam} optimizer and the \emph{Binary Cross-Entropy} loss function to compile the models. The training and test loss values are tabulated in Table \ref{train_test_loss_values}.

\noindent\textbf{Data Characterization and Classification:} For image classification we use \emph{DenseNet} \cite{huang2017densely} as our base classification model. PIs alone are not as powerful as features learnt using deep learning frameworks for image classification. However, past research works have shown topological features to carry complementary information that can be exploited to improve the overall performance of a machine learning model \cite{Dey2017,li2014persistence,som2018perturbation}. We too show results using \emph{DenseNet} in conjunction with PIs that are generated using traditional filtration techniques and using the proposed \emph{Image PI-Net} model. Figure \ref{fusion_illustration} illustrates how we pass the computed PIs as a secondary input to the base classification network. Our \emph{DenseNet} model has the following specifications: depth = $16$, number of dense blocks = $4$, number of convolution filters = $16$, growth rate = $12$, dropout rate = $0.2$ and weight decay = $10^{-4}$. We pass the generated PIs through a single 2D convolution layer with $32$ filters. This is followed by a global-average-pool layer which results in a $32$-dimensional feature vector. This feature vector is concatenated with the output of the global-average-pool layer (penultimate layer) of the \emph{DenseNet} model.

\begin{table}[t!]
	\centering
	\scalebox{0.675}{
	\begin{tabular}{ |c|c|c|c|c|} 
			\hline
			\multirow{2}{*}{ \textbf{Method}} & \multicolumn{2}{c|}{\textbf{CIFAR10}} & \multicolumn{2}{c|}{\textbf{SVHN}}\\ 
			\cline{2-5}
			& \textbf{Mean$\pm$SD} &  \textbf{p-Value} & \textbf{Mean$\pm$SD} & \textbf{p-Value} \\
			
			\hline
			DenseNet & 83.80$\pm$0.12 & - & 95.65$\pm$0.00 & - \\
			
			\textbf{DenseNet + PI} & \textbf{84.37$\pm$0.21} & 0.0153 & \textbf{95.86$\pm$0.01} & $<$0.0001  \\
			
			\textbf{DenseNet + Image PI-Net} & \textbf{84.82$\pm$0.19} & 0.0160 & \textbf{95.95$\pm$0.08} & 0.0038  \\
			
			\textbf{DenseNet + Image PI-Net FA} & \textbf{84.69$\pm$0.38} & 0.0195 & \textbf{95.84$\pm$0.06} & 0.0063 \\
			
			\textbf{DenseNet + Image PI-Net FS} & \textbf{84.59$\pm$0.17} & 0.0032 & \textbf{95.95$\pm$0.07} & 0.0020 \\
			\hline
	\end{tabular}}
	\vspace{0.05in}
	\caption{Image classification accuracy results for CIFAR10 and SVHN datasets, with the mean $\pm$ std values calculated over three runs. $P$-value is calculated with respect to the base DenseNet model.}\label{image_classification_table}
	\vspace{-0.15in}
\end{table}

\begin{figure*}[t!]
	\centering
	\includegraphics[width=0.95\linewidth]{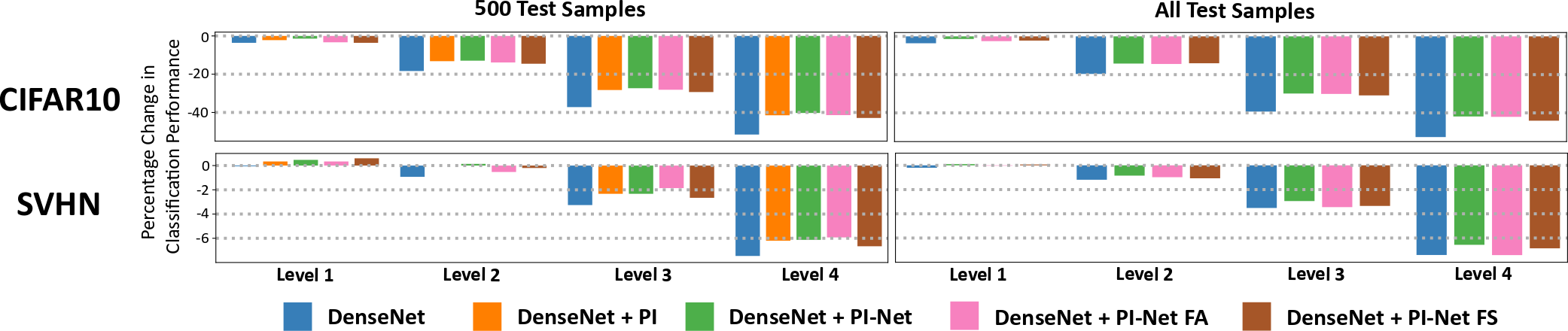}
	\caption{Percentage point drop in the classification performance on CIFAR10 (top) and SVHN (bottom) as the Gaussian noise severity increases. The percentage drop is calculated with respect to the classification performance of the DenseNet model in the absence of any Gaussian noise. Without noise, the DenseNet classification performance for 500 Test Samples and All Test Samples for CIFAR10 is $82.93\%$ and $83.80\%$, and for SVHN is $96.06\%$ and $95.65\%$. While the performance of all models drop as degradation increases, the drop of topological fusion models is less compared to just the DenseNet model.  Note, the $y$-axis is scaled different for each dataset.}\label{cifar_svhn_noise_trends}
	\vspace{-0.1in}
\end{figure*}

The classification results are averaged over three runs and are tabulated in Table \ref{image_classification_table}. We see that fusing PI feature helps improve the overall classification result for the base model on both datasets. PIs generated using the traditional filtration method and the proposed \emph{Image PI-Net} framework achieve similar results. Also, \emph{Image PI-Net FS} being trained on just 500 samples per class, achieves a classification result that is comparable to the other \emph{Image PI-Net} variants. This is useful in cases where there is limited training data for the target task. To check the significance of the different fusion cases we calculate the $P$-value for each case with respect to just the \emph{DenseNet} model. $P$-value is the area of the two-sided $t$-distribution that falls outside $\pm t$. We consistently observe a $P$-value of less than $0.05$ across all fusion cases. While we only observe marginal improvement in terms of classification accuracy, the advantage of using \emph{PI-Net} with the base classification model is made apparent in the next section.

\subsection{Robustness to Gaussian Noise in Images}

While data augmentation can help neural networks learn different transforms, TDA methods have the ability to encode different invariances by default. This could help reduce if not completely remove the need for different data variations during the training process. Here we evaluate the robustness of the different \emph{DenseNet} classification models when the test-set images are subjected to Gaussian noise. Note, the classification models were trained using the original training-set images and no data-augmentation was done during the training process. All images were first normalized to lie between $[0, 1]$. For both datasets we apply a zero-mean Gaussian noise and vary the standard deviation to the following levels: $0.02, 0.04, 0.06, 0.08$. After applying Gaussian noise we clip the pixel values in the image to lie between $[0,1]$. We refer to the four increasing severity levels of Gaussian noise as \textbf{Level 1}, \textbf{Level 2}, \textbf{Level 3}, \textbf{Level 4} or in short \textbf{L1}, \textbf{L2}, \textbf{L3}, \textbf{L4}.

Since computing PDs and PIs using traditional analytic methods is computationally expensive, we were not able to evaluate the \emph{DenseNet + PI} case on all test images. To give some perspective, computing PIs for each severity level on the test-set would take about 10 hours and 24 hours for \emph{CIFAR10} and \emph{SVHN} respectively. More information about the computational complexity is discussed in Section \ref{computation_time_section}. To compare all methods we randomly select 500 images from the test set and compare the classification performance. Figure \ref{cifar_svhn_noise_trends} shows the percentage change in the classification performance with respect to the \emph{DenseNet} method in the absence of any Gaussian noise. The effect of Gaussian noise is different for each dataset due to which the $y$-axis is scaled differently. From the bar-plots we see that the overall classification performance decreases as the severity level increases. However, the percentage decrease for \emph{DenseNet + PI} and the different \emph{DenseNet + Image PI-Net} variants is less compared to \emph{DenseNet} alone. Fusing PIs with the \emph{DenseNet} model helps incorporate robustness to different Gaussian noise. We see similar trends between the \emph{500 Test Samples} and \emph{All Test Samples} cases.

\subsection{Computation Time to Generate PIs} \label{computation_time_section}

We used the NVIDIA GeForce GTX Titan Xp graphic card with 12GB memory to train and evaluate all deep learning models. All other tasks were carried out on a standard Intel i7 CPU using Python with a working memory of 32GB. We use the Scikit-TDA software to compute PDs and PIs \cite{scikittda2019}. Table \ref{time_computation} shows the average time taken to extract PI for one image by conventional TDA methods using one CPU and the proposed \emph{PI-Net} framework on both a CPU and a GPU. The average is computed over all training images in each dataset. Using the \emph{Image PI-Net} model on a GPU, we see an effective speed up of three orders of magnitude in the computation time. Also, \emph{Image PI-Net} implemented on a CPU is still faster than the analytic method by an order of magnitude. Using a GPU we also check the time taken to compute PIs when the entire training set is passed into \emph{Image PI-Net} as a single batch. It took about 9.77$\pm$0.08 seconds for \emph{CIFAR10} and 12.93$\pm$0.05 seconds for \emph{SVHN}. This is a fraction of the time compared to the time it takes using conventional TDA tools. So far it had been impossible to compute PIs at real-time using conventional TDA approaches. However, the proposed framework allows us to easily compute PIs in real-time thereby opening doors to new real-time applications for TDA. 

 \begin{table}[t!]
	\centering
	\scalebox{0.825}{\begin{tabular}{ |c|c|c|c| } 
			\hline
			\multirow{3}{*}{ \textbf{Method}} & \multicolumn{2}{c|}{\textbf{Time ($\mathbf{10^{-3}}$ seconds)}} \\ 
			\cline{2-3}
			& \makecell{\textbf{CIFAR10}\\ \textbf{(50000 images)}}  & \makecell{\textbf{SVHN}\\ \textbf{(73257 images)}} \\
			\hline
			Conventional TDA - CPU & 3567.29$\pm$867.74  & 3433.06$\pm$622.21 \\
			\textbf{PI-Net - CPU} & \textbf{125.45$\pm$5.30} & \textbf{125.49$\pm$5.34} \\
			\textbf{PI-Net - GPU} & \textbf{2.52$\pm$0.02} & \textbf{2.19$\pm$0.02} \\
			\hline
	\end{tabular}}
	\vspace{0.05in}
	\caption{Comparison of the average time taken to compute PIs for one image using conventional TDA tools and the proposed PI-Net model. The time reported is averaged over all images present in the training set of each dataset.}\label{time_computation}
	\vspace{-0.325in}.
\end{table}

\vspace{-0.05in}
\section{Conclusion and Future Work}\label{conclusion_future_work}
\vspace{-0.05in}

In this paper we took the first step in using deep learning to extract topological feature representations. We developed a simple, effective and differentiable architecture called to extract PIs directly from time-series and image data. \emph{PI-Net} has a significantly lower computational complexity compared to using conventional topological tools. We show improvements in classification performance on two accelerometer and two image datasets. Despite observing marginal improvement in image classification accuracy, the benefit of using \emph{PI-Net} with the base classification network is made apparent through the robustness to Gaussian noise experiment. 

For future work we would like to explore more sophisticated deep learning architectures that can allow us to learn mappings between higher dimensional data and their corresponding topological feature representations. We would also like to see how deep learning can be further used to generate other kinds of topological representations and test their robustness to different image deformations like contrast, blur and affine transformations. 

{\small
\bibliographystyle{ieee_fullname}
\bibliography{egbib}
}

\end{document}